\documentclass{article}

\usepackage{arxiv}

\usepackage[utf8]{inputenc} 
\usepackage[T1]{fontenc}    
\usepackage{hyperref}       
\usepackage{url}            
\usepackage{booktabs}       
\usepackage{amsfonts}       
\usepackage{nicefrac}       
\usepackage{microtype}      
\usepackage{lipsum}
\usepackage{graphicx}
\usepackage{amsmath}
\usepackage{multirow}

\title{Transferred Fusion Learning using Skipped Networks}

\author{
  Vinayaka R Kamath \\
  Dept. of CSE \\
  PES University\\ Bengaluru, India \\
  \texttt{vinayakrkamath@pesu.pes.edu}
   \And
 Vishal S Rao \\
  Dept. of CSE \\
  PES University\\ Bengaluru, India \\
  \texttt{vishalmkb@gmail.com}
  \And 
  Varun M \\
  Dept. of CSE \\
  PES University\\ Bengaluru, India \\
  \texttt{varunm1602@gmail.com}
}

\begin{document}
\maketitle

\begin{abstract}
Identification of an entity that is of interest is prominent in any intelligent system. The visual intelligence of the model is enhanced when the capability of recognition is added. Several methods such as transfer learning and zero shot learning help to reuse the existing models or augment the existing model to achieve improved performance at the task of object recognition. Transferred fusion learning  is one such mechanism that intends to use the best of both worlds and build a model that is capable of outperforming the  models involved in the system. We propose a novel mechanism to amplify the process of transfer learning by introducing a student architecture where the networks learn from each other.
\end{abstract}


\section{Introduction}
Deep learning is a specialization of machine learning which is derived from the neurological structure of the human brain. A main component present in the neurological structure of the brain is the neuron. The neuron consists of axons which serves as a medium to transfer electrical impulses. Dendrites are used to receive information from other neurons in the system. Soma is the cell body  which defines the neuron’s structure. Similarly in case of deep learning, the nodes present in the layers serve the functionality of soma, the links connecting the nodes are served by axons and dendrites. The nodes are arranged in a sequential manner, one on top of the other. A layer for the neural network is formed by stacking neurons on top of one another. A neural network consists of three layers, input, hidden and output. These three layers are connected via a link where one layer passes the output to the other layer. On each link, the trained weights decide the weightage of the output. Initially, random weights are assigned to all the neurons in the network . We pass a training dataset composed of inputs and the actual target output. The neural network predicts the targeted output from the weights initially assigned and the error function takes the difference between the actual and the predicted output. This error is then backpropagated to re-adjust the weights based on the error.

\begin{figure}
    \centering
    \includegraphics[width=\textwidth]{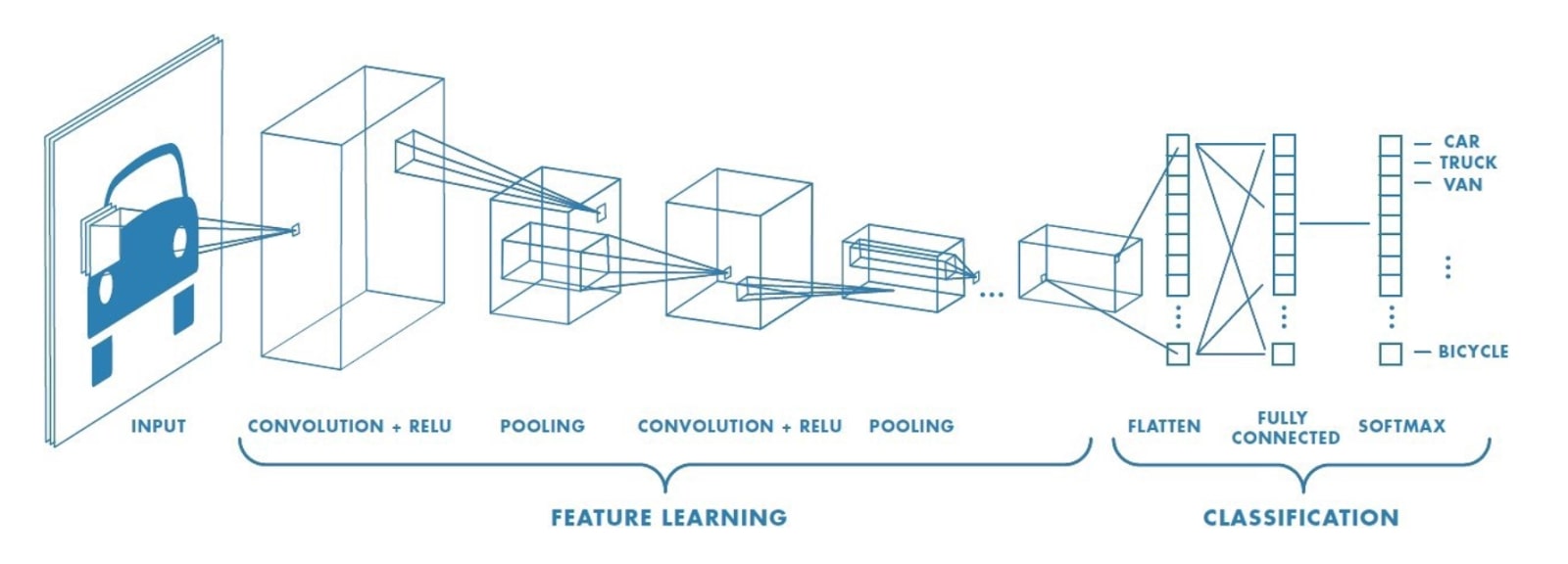}
    \caption{General Architecture used for Image Classification}
    \label{fig:general}
\end{figure}

Different types of layers can be cascaded to achieve the intended purpose. The neural network can mimic a universal hypothesis that is proficient in approximating the true function given a suitable architecture. A deep enough neural network can perform feature extraction in the starting stages of the network and learn to associate the features with the desired task in the latter stages of the network. Introducing non-linearlity in the architecture of the model can help the model learn more complex features. Choosing appropriate hyper-parameters can enhance the capabilities of the model, electing a challenging enough dataset can augment the training process.

The trained network is capable of performing numerous tasks that offer intelligence to the system that is using it. In one sense it aids in replicating the behaviour of an intelligent agent. These networks can identify objects, rectify distortions, enhance the features as well as remove imperfections. The necessity of designing an optimal architecture to achieve the desired goal is of paramount. We propose one such architecture that is well versed in the process of image classification and completely utilizes the subsystems i.e other smaller neural networks in our case, even the ones that are prone to underfitting.


The components constituting the suggested approach and the architecture is presented over the next sections. Section \ref{sec:essentials} describes the ideas that over lap with the advocated approach and form the building blocks to the recommended architecture. Section \ref{sec:experimentation} briefs about the experimentation procedure, the datasets used to prove the correctness and describes the architecture itself. The last section presents an analysis over the acquired results and further concludes with the possible improvements and future scope.

\section{Essentials of the proposed architecture}
\label{sec:essentials}
The proposed system excels at correlating the features learned by it constituents. The following concepts overlap with the design proposed in our work and describe the basic idea that is required to prove the correctness of the new model.

\subsection{Convolutional Neural Network}

Convolutional neural networks~\cite{cnn} take an image as an input and predict the type of the image output class from the classes provided as a part of the training set~\cite{cnn_image}. The output classes are the labels assigned for the training set of images. The layers present in the network serve different purposes. An input image which lies in either RGB, HSV, CMYK colour space is fed to the convolution layer~\cite{cnn_hsv}. A convolutional layer extracts the distinguishing features from the image like the outer edges, shades of the spatial context, uniformity in the patterns and lighting intensities. A kernel, which is a matrix smaller than the input image moves through the image based on the stride value $S$~\cite{dl_book}. $S$ defines the number of pixels the kernel matrix needs to move when it inspects through the input image. The size of the input matrix is increased by padding it by a padding depth. We acquire a feature map or activation map of reduced dimensions from the input image. The size of an image of order $W_1 * H_1 * D_1$ after convolution is given by:
\begin{equation}
    \begin{split}
        W_2 & = \frac{W_1 - F + 2 \cdot P}{S} + 1  \\
        H_2 & = \frac{H_1 - F + 2 \cdot P}{S} + 1 \\
        D_2 & = D_1 \\
    \end{split}
\end{equation}
where $F$ is the size of the kernel, $P$ denotes padding size or padding depth and $S$ is the stride value for a output feature map of dimension $W_2 * H_2 * D_2$. 

These feature maps are passed as an input to an activation function to introduce non linearity in the model~\cite{nonlinear}. Variety of activation functions are used and tested according to the requirement of the model. All the networks that form the constituents of our design use Rectified Linear Unit(ReLU)~\cite{relu}  as the activation function right after the convolutional layer. The values present in the entries of the feature map are either converted to zero or are left undisturbed based on the magnitude of the value i.e the entries that are negative are converted to zero and the positive values are left unchanged. Equation \ref{eq:relu} represents the definition of the ReLU function.

\begin{equation}
    \label{eq:relu}
    F(x) = max(0, Z)
\end{equation}
where $Z$ can be linear or quadratic in nature. A pooling operation is applied on the activated feature map to reduce the computational complexity of the model. The units in the activated feature maps are dropped based on a specific criteria~\cite{pooling}. A max pooling is applied to the activated feature map in the pooling layer in every stage of the proposed architecture. In max pooling, a suitable matrix size that is lesser than the feature map is chosen. The matrix from the pooling layer moves across the map based on a stride value $S$ and the maximum of the values seen by this matrix is computed. A new matrix with reduced dimensions is obtained after the procedure. The size of the output matrix can be computed using the equation \ref{eq:pool}, given the input matrix is of the order $ W_1 * H_1 * D_1$.
\begin{equation}
    \label{eq:pool}
    \begin{split}
        W_2 & = \frac{W_1 - F}{S} + 1  \\
        H_2 & = \frac{H_1 - F}{S} + 1 \\
        D_2 & = D_1 \\
    \end{split}
\end{equation}
where $F$ is the size of the pooling kernel, $S$ is the stride value that influences the shift in the pooling window on the output feature map of dimension $W_2 * H_2 * D_2$.

The max pooled matrix is converted to a single vector by the process of flattening, this is done by stacking the succeeding rows side by side. This flattened matrix is fed as an input to the fully connected layer. The fully connected layer is a feed forward artificial neural network that performs backpropagation~\cite{back} to every instance of the training example. After training the model for a good number of epochs, the model is able to classify an image from the test set not present as a part of the training set.

\subsection{Transfer Learning}

Whenever humans encounter a new task to fulfill, they use their  learning experience from previous tasks instead of learning everything from scratch. Traditional learning methods need to train from scratch based on specific problems~\cite{tl_survey}. Transfer learning~\cite{tl_book} is an approach where a model when presented with a new problem applies its experiences from the previous training instances. The need of transfer learning can be explained with an example. When task t is presented where we need to detect a list of objects present in images of a restobar. The traditional learning model would train only on the set of restobar images and hence it's domain is confined to the use case of a restobar. Suppose we need to extend the task of detecting objects from images previously under consideration to a new set of images such as a parks, this model is bound to perform poorly as it was trained only on a narrow set of instances,in the given example, that being the images of a restobar. The process of training a model is an added burden. Since not enough data is available to train the model on the park images, the model is unable to excel at this task. Transfer learning can aid in accelerating this process and land the model in superior terminating states than the one that is trained on a new dataset~\cite{tl_collection}.

\begin{figure}[bht!]
    \centering
    \includegraphics[width=0.6\textwidth]{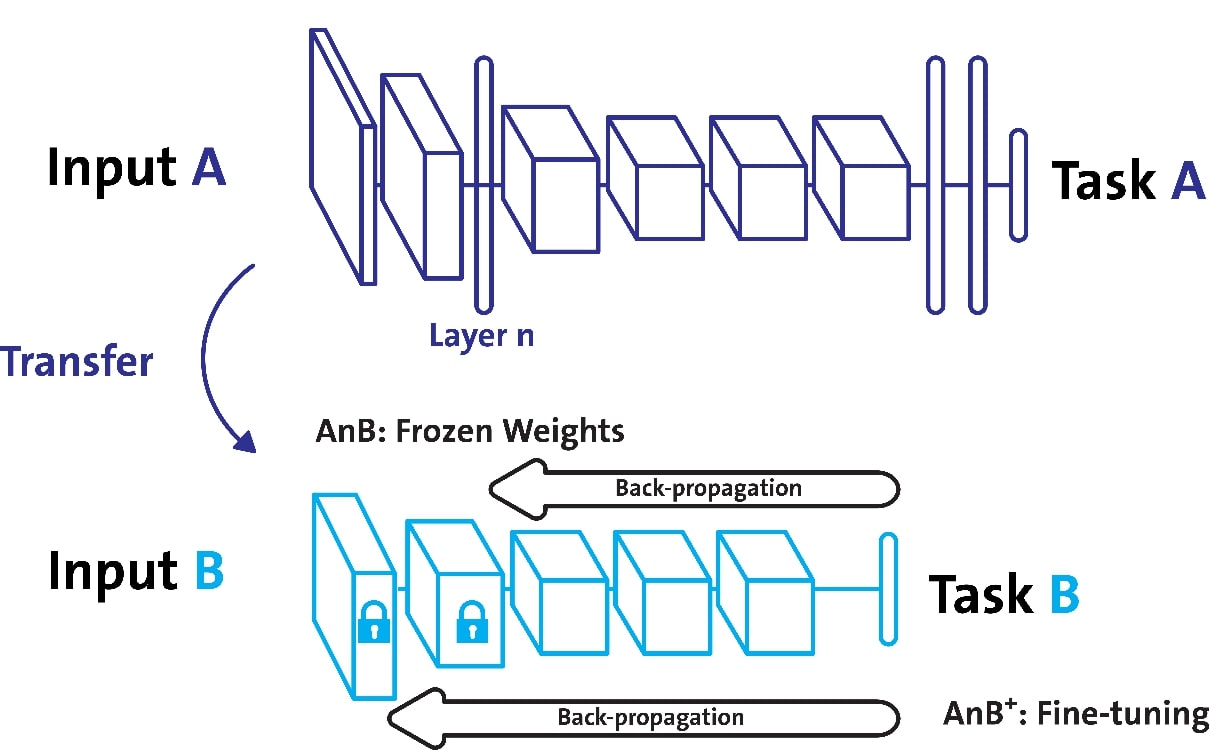}
    \caption{Illustration of Transfer Learning}
    \label{fig:tl}
\end{figure}

Transfer learning can be described using two primal components:
\begin{itemize}
    \item Domain $D = (Q, P(X))$
    \item[] where $Q$ represents the feature space and $P(X)$ is the marginal distribution given by $X = \{x_1, x_2, ..., x_n\}$ where $x_i$ belongs to $Q$.
    \item Task $T = (R, P(Y/X)) = (R, n)$ 
    \item[] where $Y$ is given by $Y = \{y_1, y_2, ..., y_n\}$ and the label space is denoted by $R$ and $y_i \in R$. $n$ is a predictive function computed by training on the pair $(x_i, y_i)$ where $x_i \in Q$ and $y_i \in R$.
\end{itemize}
The output for every feature vector in the domain is computed using  $n(x_i) = y_i$. The model is presented with a source domain and the corresponding target domain represented using $D_s$ and $T_s$ respectively, along with the target domain and the target task denoted by $D_t$ and $T_t$ respectively. This systems enables the  model to learn $P( Y_t | X_t )$ present in the $D_t$, using the trained experience it gained from $D_s$ and $T_s$ under the condition that $D_s != D_t$ and $ T_s != T_t$.

The process of transfer learning can be mainly categorized into three methods:
\begin{itemize}
    \item Inductive transfer learning: The constraints  $ D_s = D_t $ and $ T_s = T_t $ has to be followed, but the tasks of the source and target are remarkably different from each other. The tasks in the source and target domains are supervised~\cite{tl_inductive}.
    \item Unsupervised transfer learning~\cite{tl_unsupervised} : The constraints of the inductive transfer learning are followed while the main focus is on unsupervised tasks in the target domain.
    \item Transductive transfer learning~\cite{tl_transductive}: The domains and labels do not exhibit any overlap i.e $D_s != D_t$ and $T_s != T_t$, but the tasks of both the source and target domain are similar intuitively.    
\end{itemize}

\subsection{Skip Connections}

\begin{figure}[h!]
    \centering
    \includegraphics[width=0.8\textwidth]{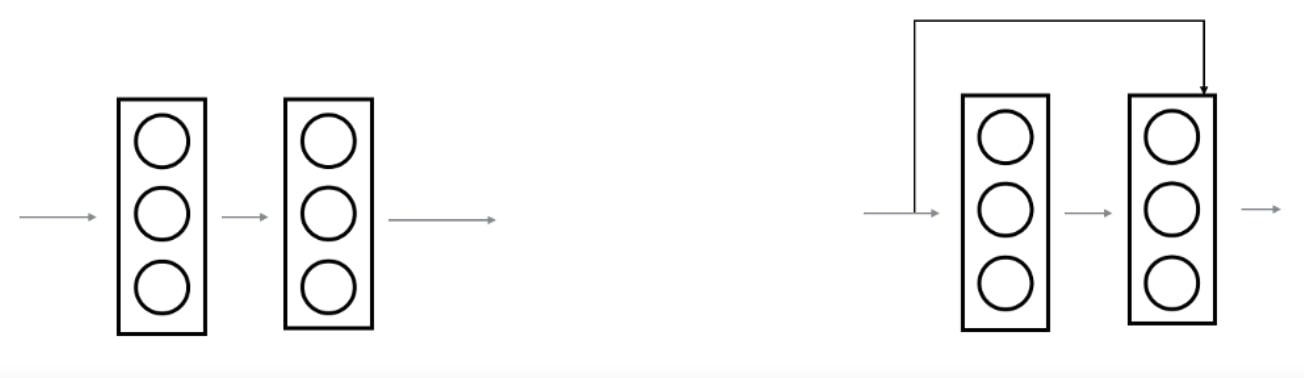}
    \caption{Representation of how skip connections enable the model to carry forward the feature map}
    \label{fig:my_label}
\end{figure}

Initially, training extremely deep neural networks posed a major problem due to the gradient vanishing phenomenon~\cite{vanishing} encountered during the backpropagation~\cite{back} stage of the training process. This was a major setback, especially when the network was extremely deep~\cite{going_deeper}. As a result of the vanishing gradient problem, the deeper networks performed worse than the shallower networks in a lot of cases. In order to combat this issue skip connections were introduced~\cite{skip_connections}. A new neural network architecture known as ResNet~\cite{resnet} came up with the idea of skip connections. These connections are between the input of the convolution layer and its output. The dimensions of the skipped ends were to be ensured to agree with one another. The dimensions of the activation maps at the ends of the skip connections were to be adjusted to a common value using a convolutional block if they did not agree with one another~\cite{res_impact}. The new design worked for the following reasons:
\begin{enumerate}
    \item There is another path from the input to the output of a convolution layer(that doesn't involve a convolution operation) which is responsible for reduction in gradient loss during the backpropagation phase. This prevents the gradient from vanishing during propagation.
    \item The identity function present makes sure that the higher layer will perform at least as good as the lower layers.
\end{enumerate}

\section{Experimentation}
\label{sec:experimentation}
\subsection{Datasets}
\subsubsection{Natural Images}
\begin{table}[tbh!]
    \centering
    \caption{Image count of each class}
    \label{tab:count}
    \begin{tabular}{|l|l|l|l|l|l|l|l|l|}
        \hline
         { \bfseries Class} & Airplane & Car & Motorbike & Flower & Fruit & Person & Cat & Dog  \\
         \hline
         { \bfseries Count} & 727 & 968 & 788 & 843 & 1000 & 986 & 885 & 702 \\
         \hline
    \end{tabular}
\end{table}
The dataset consists of the images of different objects that are captured in the wild~\cite{natural_images}. There are a total of 8 different classes of images that exhibit enough variations for the model to understand the features. The class labels include airplane, car, motorbike, flower, fruit, person, cat and dog. The dimensions of the images varied from class to class ranging from 100x100 to as big as 700x400. 6899 images were used for training and validating. The dataset posed challenges in orientation, lighting conditions and temporal inconsistencies of the objects. The distribution of the dataset can be observed by looking at table \ref{tab:count}.
\begin{figure}[htb!]
    \centering
    \fbox{\includegraphics[width=0.6\textwidth]{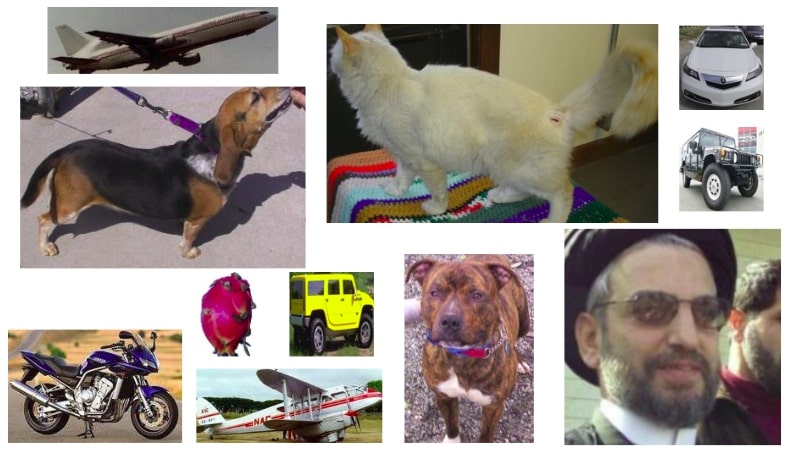}}
    \caption{Sample Images from Natural Images Dataset}
    \label{fig:natural_images}
\end{figure}
\subsubsection{MNIST}

The dataset~\cite{mnist} consists of 70,000 hand written digits which are monochromatic. It has 10 classes, each per digit and offers variations in translation, skewness, region of interest and the hand writing followed in the image. Each image is of the dimension 28x28 and each instance can be uniquely identified by the characteristics it exhibits. The dataset does not pose any serious challenge and can be considered as a benchmark to prove the correctness of the algorithm.
\subsubsection{CIFAR-100}
Closely resembling the CIFAR-10, the dataset provides images of dimension 32x32 distributed over a variety of classes that have little to no overlap. The dataset has exactly 100 classes and provides 600 images per class. The images are taken from different angles from the object of interest and offers higher level of challenge compared to the other two datasets. There is significant variations in the lighting conditions, angle of the object and even the object of interest is occluded in few cases.
\begin{figure}[thb!]
    \centering
    \includegraphics[width=0.8\textwidth]{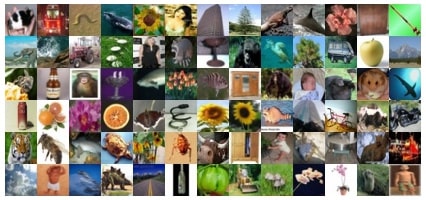}
    \caption{Variations exhibited by CIFAR-100 Dataset}
    \label{fig:cifar}
\end{figure}
\subsection{Architecture}

\subsubsection{Preprocessing}
The images in the dataset were of different dimensions, this introduced the necessity to resize the images into the same size. To make them uniform, all the images were resized to the dimension 100x100 as this dimension introduced minimal distortions when compared to other dimensions that were manually inspected. The dataset was divided into training set and testing (validation) set  with training set having 80\% of the dataset. 

\subsubsection{Proposed Methodology}

The new model intends to introduce skip connections in between the pre-trained models. The output of a block, consisting of a convolutional layer and an activation unit is fed to the other model in the system. The outflow of the models are exchanged to make sure that the other model learns the features of its counterpart. Dense layers are appended to the system after making sure that the outputs of the models are concatenated. The system is trained using a small set of data and the weights are updated to make sure the desired output is obtained.

\begin{figure}[bht!]
    \centering
    \includegraphics[scale=0.35]{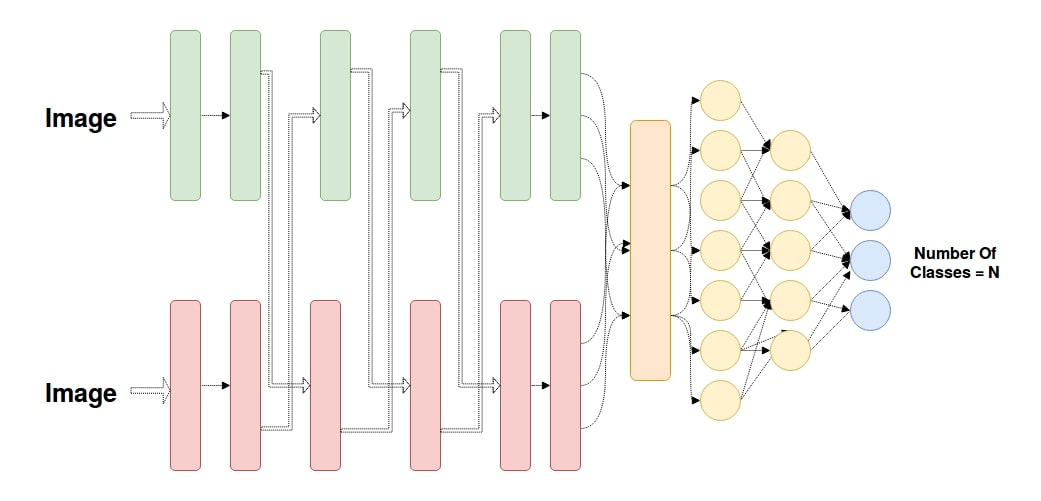}
    \caption{Schematic of the Proposed Architecture}
    \label{fig:arch}
\end{figure}

The existing architecture of two pre-trained models is trained on a dataset that is completely different from the hybrid model. The new architecture learns to re-use the features learnt from the previous datasets and learns the feature map accordingly. The new model is not only learning the features from the data provided to it, but also learns the activation patterns from the other model. The models are co-learning from each other as well as from the data provided during the secondary training. The models are inspected for matching dimensions or dimensions that are almost similar. In case the dimensions exactly match, the output from the first model is added to the output from the counterpart by directly feeding it to  the input of the next layer. The same procedure is repeated for the other model in the system. In layers where the dimensions slightly overlap don't fit perfectly, a new convolution block is added to resize the activation map~\cite{resize} to the desired shape. Since the dimensions match after the convolution operation, the same procedure is followed there after.

The skip connections helps to transfer the feature map between the models~\cite{skip_path}. They act like a bridge that helps to transfer the features from one model to another. The model learn to correlate the patterns in the input and the activation pattern from the other model in the system. This enhances the operation of transfer learning, where not only the features from the old model is transferred to the new model, features from a model with similar architecture is also transferred to the hybrid model. This enhances the process of retraining the model with a smaller dataset, as there is much more for the new model to learn.

\section{Results and Conclusion}
\label{sec:results}
The proposed approach was tested on a multitude of combinations to prove the goodness of the new architecture. The datasets were juggled for the same pair to thoroughly critique the proposed model. Inception V3~\cite{inception} and SqueezeNet~\cite{squeezenet} are widely popular pretrained networks that are equally versatile and exhibit remarkable results with transfer learning. SqueezeNet is composed of a light weight architecture where as the inception is more deeper and dense, both are very different from one another and are appropriate to provide insights into the suggested architecture.

A CNN with 16 layers was designed and trained to analyse the algorithm accordingly~\cite{xception}. The network consisted of residual connections after every 4th block and used batch normalization to improve the understanding of the data. The network was deep enough to understand intricate details of data presented to it and was large enough to pair up with the deeper models such as inception and squeezenet. This custom designed neural network is referred to as "custom model" on the context here after.

Table \ref{tab:result} throws light over the experimentation performed and reports the results obtained. It can be observed that the architecture involving the pre-trained models performs much better that the architecture that use custom model as one of its constituents. This is expected as the pre-trained models are trained on datasets such as ImageNet which contains millions of images. Although inception and squeezenet are far more superior than the custom model, the custom model performs exceptionally well when it is paired up with pre-trained models using the suggested architecture. For the CIFAR-100 dataset, the state-of-the-art models show ~55\% to ~70\% accuracy while the proposed architecture is right under the margin. The suggested model is a huge improvement over transfer learning as well as training a model from scratch when it comes to the CIFAR-100 dataset. The suggested architecture performs remarkably well on the MNIST dataset as well and falls right in the creamy layer. Natural images dataset is not considered a benchmark among the community, although widely used, there is little to no information on the performance of the top notch architecture on the dataset. Considering the number of classes and the challenges it poses, the proposed architecture performed exceptionally well in general. Overall, the system seem to perform satisfactory in most cases if not exceptionally well. 

\begin{table}[t!]
    \centering
    \caption{Results from Experimentation}
    \label{tab:result}
    \begin{tabular}{|l|l|l|l|l|l|}
        \hline
         Model 1 & Model 2 & Pretraining& Pretraining& Retraining & Hybrid\\
         && Datatset & Dataset & & Learning \\
         && (Model 1) & (Model 2) & Dataset & Accuracy \\
         \hline
         Inception V3 & SqueezeNet & -- & -- &  CIFAR-100 & 52.18\% \\
         Inception V3 & SqueezeNet & -- & -- &  MNIST & 97.21\% \\
         Inception V3 & SqueezeNet & -- & -- &  Natural Images & 89.57\% \\
         Inception V3 & Custom & -- & MNIST &  CIFAR-100 & 45.62\% \\
         Custom & SqueezeNet & Natural Images & -- & CIFAR-100 & 43.28\% \\
         Custom & SqueezeNet & MNIST & -- & Natural Images & 60.82\% \\
         Custom & Inception V3 & CIFAR-100 & -- & Natural Images & 73.23\% \\
         Custom 1 & Custom 2 & MNIST & CIFAR-100 & Natural Images & 62.23\% \\
         Custom 1 & Custom 2 & CIFAR-100 & CIFAR-100 & Natural Images & 68.48\% \\
         \hline
    \end{tabular}
\end{table}

To test the goodness of the suggested methodology and its superiority over the traditional transfer learning approach, the experimentation on the custom model was closely monitored. The custom model was trained on one single dataset and another model trained on a seperate dataset was combined with this model using the transferred fusion learning approach. The individual models were also subjected to transfer learning and were tested on the same dataset. Table \ref{tab:result2} briefs about the experimentation procedure followed and the results exhibited by the new approach. It can be observed that the new approach shows marginal improvements over the transfer learning approach in some cases while huge improvements over others. The suggested methodology works pretty well when the features are very complex and the network is not able to generalize the patterns thoroughly. Combining two pre-trained models enables both of them to augment its learning procedure when unseen data is presented to it during training procedure. 

\begin{table}[tbh!]
    \centering
    \caption{Improvements in custom model}
    \label{tab:result2}
    \begin{tabular}{|l|l|l|l|l|l|}
        \hline
         Pretraining & Pretraining & Retraining & TL Accuracy & TL Accuracy & Hybrid \\
         Dataset & Dataset & Dataset & (Model 1) &(Model 2) & Learning \\ 
         (Model 1)& (Model 2)& & & & Accuracy \\
         \hline
         MNIST & CIFAR-100 & Natural Images &  52.86\% & 55.15\% & 62.23\% \\
         Natural Images & MNIST & CIFAR-100 &  35.58\% & 34.29\% & 38.17\% \\
         CIFAR-100 & CIFAR-100 & Natural Images &  42.15\% & 38.46\% & 68.48\% \\
         \hline
    \end{tabular}
\end{table}

It is safe to conclude that the hybrid model performed reasonably well and can be scaled for solving real world problems in real time. The proposed architecture showed appreciable results considering the ease of implementation and the time to make it production ready.  The hybrid model outperformed the transfer learning by 10\%-15\% in several instances. It can be inferred that combining two pre-trained models using skip connections generates a better model that out performs the model trained using transfer learning in several cases and performs reasonably well in all the cases.

Although the procedure is computationally expensive compared to the traditional processes, the improved accuracy makes it suitable for several applications. The trade off between training time and accuracy is highly implementation specific. The proposed approach shows great potential to be extended to one shot learning methodology rather than simply as a tool for image classification. The model can learn to infer the differences in the features from itself and the corresponding pair. This opens new door ways to several other applications. The advocated architecture has proven to be worthy and acts as an effective counterpart to the traditional transfer learning methodology and proves that the architecture is capable enough to be extended to several applications.

\bibliographystyle{unsrt}  
\bibliography{template}  

\begin{thebibliography}{10}

\bibitem{cnn}
Alex Krizhevsky, Ilya Sutskever, and Geoffrey~E Hinton.
\newblock Imagenet classification with deep convolutional neural networks.
\newblock In {\em Advances in neural information processing systems}, pages
  1097--1105, 2012.

\bibitem{cnn_image}
Dan Cire{\c{s}}an, Ueli Meier, and J{\"u}rgen Schmidhuber.
\newblock Multi-column deep neural networks for image classification.
\newblock {\em arXiv preprint arXiv:1202.2745}, 2012.

\bibitem{cnn_hsv}
Patrice~Y Simard, David Steinkraus, John~C Platt, et~al.
\newblock Best practices for convolutional neural networks applied to visual
  document analysis.
\newblock In {\em Icdar}, volume~3, 2003.

\bibitem{dl_book}
Yann LeCun, Yoshua Bengio, and Geoffrey Hinton.
\newblock Deep learning.
\newblock {\em nature}, 521(7553):436, 2015.

\bibitem{nonlinear}
Tsung-Yu Lin, Aruni RoyChowdhury, and Subhransu Maji.
\newblock Bilinear cnn models for fine-grained visual recognition.
\newblock In {\em Proceedings of the IEEE international conference on computer
  vision}, pages 1449--1457, 2015.

\bibitem{relu}
Vinod Nair and Geoffrey~E Hinton.
\newblock Rectified linear units improve restricted boltzmann machines.
\newblock In {\em Proceedings of the 27th international conference on machine
  learning (ICML-10)}, pages 807--814, 2010.

\bibitem{pooling}
Chen-Yu Lee, Patrick~W Gallagher, and Zhuowen Tu.
\newblock Generalizing pooling functions in convolutional neural networks:
  Mixed, gated, and tree.
\newblock In {\em Artificial Intelligence and Statistics}, pages 464--472,
  2016.

\bibitem{back}
Yann LeCun, D~Touresky, G~Hinton, and T~Sejnowski.
\newblock A theoretical framework for back-propagation.
\newblock In {\em Proceedings of the 1988 connectionist models summer school},
  volume~1, pages 21--28. CMU, Pittsburgh, Pa: Morgan Kaufmann, 1988.

\bibitem{tl_survey}
Sinno~Jialin Pan and Qiang Yang.
\newblock A survey on transfer learning.
\newblock {\em IEEE Transactions on knowledge and data engineering},
  22(10):1345--1359, 2009.

\bibitem{tl_book}
Lisa Torrey and Jude Shavlik.
\newblock Transfer learning.
\newblock In {\em Handbook of research on machine learning applications and
  trends: algorithms, methods, and techniques}, pages 242--264. IGI Global,
  2010.

\bibitem{tl_collection}
Hoo-Chang Shin, Holger~R Roth, Mingchen Gao, Le~Lu, Ziyue Xu, Isabella Nogues,
  Jianhua Yao, Daniel Mollura, and Ronald~M Summers.
\newblock Deep convolutional neural networks for computer-aided detection: Cnn
  architectures, dataset characteristics and transfer learning.
\newblock {\em IEEE transactions on medical imaging}, 35(5):1285--1298, 2016.

\bibitem{tl_inductive}
Alexandru Niculescu-Mizil and Rich Caruana.
\newblock Inductive transfer for bayesian network structure learning.
\newblock In {\em Artificial intelligence and statistics}, pages 339--346,
  2007.

\bibitem{tl_unsupervised}
Rajat Raina, Alexis Battle, Honglak Lee, Benjamin Packer, and Andrew~Y Ng.
\newblock Self-taught learning: transfer learning from unlabeled data.
\newblock In {\em Proceedings of the 24th international conference on Machine
  learning}, pages 759--766. ACM, 2007.

\bibitem{tl_transductive}
Andrew Arnold, Ramesh Nallapati, and William~W Cohen.
\newblock A comparative study of methods for transductive transfer learning.
\newblock In {\em ICDM Workshops}, pages 77--82, 2007.

\bibitem{vanishing}
Sepp Hochreiter.
\newblock The vanishing gradient problem during learning recurrent neural nets
  and problem solutions.
\newblock {\em International Journal of Uncertainty, Fuzziness and
  Knowledge-Based Systems}, 6(02):107--116, 1998.

\bibitem{going_deeper}
Christian Szegedy, Wei Liu, Yangqing Jia, Pierre Sermanet, Scott Reed, Dragomir
  Anguelov, Dumitru Erhan, Vincent Vanhoucke, and Andrew Rabinovich.
\newblock Going deeper with convolutions.
\newblock In {\em Proceedings of the IEEE conference on computer vision and
  pattern recognition}, pages 1--9, 2015.

\bibitem{skip_connections}
Xiaojiao Mao, Chunhua Shen, and Yu-Bin Yang.
\newblock Image restoration using very deep convolutional encoder-decoder
  networks with symmetric skip connections.
\newblock In {\em Advances in neural information processing systems}, pages
  2802--2810, 2016.

\bibitem{resnet}
Zifeng Wu, Chunhua Shen, and Anton Van Den~Hengel.
\newblock Wider or deeper: Revisiting the resnet model for visual recognition.
\newblock {\em Pattern Recognition}, 90:119--133, 2019.

\bibitem{res_impact}
Christian Szegedy, Sergey Ioffe, Vincent Vanhoucke, and Alexander~A Alemi.
\newblock Inception-v4, inception-resnet and the impact of residual connections
  on learning.
\newblock In {\em Thirty-First AAAI Conference on Artificial Intelligence},
  2017.

\bibitem{natural_images}
Dengsheng Lu and Qihao Weng.
\newblock A survey of image classification methods and techniques for improving
  classification performance.
\newblock {\em International journal of Remote sensing}, 28(5):823--870, 2007.

\bibitem{mnist}
Yann LeCun.
\newblock The mnist database of handwritten digits.
\newblock {\em http://yann. lecun. com/exdb/mnist/}, 1998.

\bibitem{resize}
HyunWook Park, YoungSeo Park, and Seung-Kyun Oh.
\newblock L/m-fold image resizing in block-dct domain using symmetric
  convolution.
\newblock {\em IEEE Transactions on Image Processing}, 12(9):1016--1034, 2003.

\bibitem{skip_path}
Jin Yamanaka, Shigesumi Kuwashima, and Takio Kurita.
\newblock Fast and accurate image super resolution by deep cnn with skip
  connection and network in network.
\newblock In {\em International Conference on Neural Information Processing},
  pages 217--225. Springer, 2017.

\bibitem{inception}
Christian Szegedy, Vincent Vanhoucke, Sergey Ioffe, Jon Shlens, and Zbigniew
  Wojna.
\newblock Rethinking the inception architecture for computer vision.
\newblock In {\em Proceedings of the IEEE conference on computer vision and
  pattern recognition}, pages 2818--2826, 2016.

\bibitem{squeezenet}
Forrest~N Iandola, Song Han, Matthew~W Moskewicz, Khalid Ashraf, William~J
  Dally, and Kurt Keutzer.
\newblock Squeezenet: Alexnet-level accuracy with 50x fewer parameters and< 0.5
  mb model size.
\newblock {\em arXiv preprint arXiv:1602.07360}, 2016.

\bibitem{xception}
Fran{\c{c}}ois Chollet.
\newblock Xception: Deep learning with depthwise separable convolutions.
\newblock In {\em Proceedings of the IEEE conference on computer vision and
  pattern recognition}, pages 1251--1258, 2017.

\end{thebibliography}

\end{document}